\documentclass[conference]{IEEEtran}
\IEEEoverridecommandlockouts
\usepackage{cite}
\usepackage{amsmath,amssymb,amsfonts}
\usepackage{algorithmic}
\usepackage{graphicx}
\usepackage{textcomp}
\usepackage{xcolor}
\usepackage{subfigure}
\def\BibTeX{{\rm B\kern-.05em{\sc i\kern-.025em b}\kern-.08em
    T\kern-.1667em\lower.7ex\hbox{E}\kern-.125emX}}
\begin{document}

\title{Are Explainability Tools Gender Biased? A Case Study on Face Presentation Attack Detection
\thanks{This research work has been funded by the German Federal Ministry of Education and Research and the Hessian Ministry of Higher Education, Research, Science and the Arts within their joint support of the National Research Center for Applied Cybersecurity ATHENE. This work has been partially funded by the German Federal Ministry of Education and Research through the Software Campus Project.}}

\author{\parbox{16cm}{\centering
    {\large Marco Huber$^{1}{}^{,}{}^{2}$, Meiling Fang$^{1}{}^{,}{}^{2}$, Fadi Boutros$^{1}{}$, Naser Damer$^{1}$ \\
    {\normalsize
    $^1$Fraunhofer Institute for Computer Graphics Research IGD, Darmstadt, Germany\\
    $^2$Department of Computer Science, TU Darmstadt, Darmstadt, Germany
    }}}}

\maketitle

\begin{abstract}
Face recognition (FR) systems continue to spread in our daily lives with an increasing demand for higher explainability and interpretability of FR systems that are mainly based on deep learning. While bias across demographic groups in FR systems has already been studied, the bias of explainability tools has not yet been investigated. As such tools aim at steering further development and enabling a better understanding of computer vision problems, the possible existence of bias in their outcome can lead to a chain of biased decisions. In this paper, we explore the existence of bias in the outcome of explainability tools by investigating the use case of face presentation attack detection.  By utilizing two different explainability tools on models with different levels of bias, we investigate the bias in the outcome of such tools. Our study shows that these tools show clear signs of gender bias in the quality of their explanations.
\end{abstract}


\section{Introduction} 
\label{sec:intro}
Face recognition (FR) is increasingly present in our everyday lives, whether it is crossing borders or unlocking our smartphones. 
Current FR systems \cite{DBLP:conf/cvpr/BoutrosDKK22, DBLP:conf/cvpr/MengZH021} achieve outstanding performances that can even exceed those of humans \cite{DBLP:journals/ivc/PhillipsO14}, but are difficult for humans to understand and analyse due to the opacity of the deep learning methods used \cite{DBLP:journals/corr/abs-2208-09500}. 
To increase the understanding of the deep learning models' performance and their behavior in computer vision tasks, several explainability methods have been proposed, such as GradCAM \cite{DBLP:conf/iccv/SelvarajuCDVPB17} or GradCAM++ \cite{DBLP:conf/wacv/ChattopadhyaySH18}, to highlight important areas for a given task on an image. These methods are gaining increasing attention in the field of biometrics \cite{DBLP:journals/iet-bmt/SequeiraGSP021, DBLP:conf/wacv/NetoSC22, DBLP:journals/corr/abs-2208-09500}. Explainability tools aim at enhancing trust in biometrics technologies and can also lead to new solutions for challenges facing biometric systems, such as differential performance and bias. Bias refers to relative performance differences towards certain demographic or non-demographic subgroups \cite{DBLP:journals/corr/abs-2103-01592} that might enable unfair behavior of the system or systematic discrimination. 

While recent works investigated the demographic bias and the fairness of FR \cite{DBLP:journals/tbbis/PereiraM22, DBLP:journals/corr/abs-2103-01592} and its related tasks such as face image quality \cite{9909867, DBLP:conf/icb/TerhorstKDKK20} and face presentation attack detection \cite{DBLP:journals/corr/abs-2209-09035, DBLP:journals/fi/AlshareefYRA21}, the bias that might be present in the explanations provided by the emerging explainability tools used to increase the interpretability of the used models has not been investigated so far. 
Different explainability tools have already been used in biometric systems \cite{DBLP:journals/corr/abs-2110-11001, DBLP:conf/wacv/NetoSC22, DBLP:conf/fgr/MirzaalianHSMA21}, mainly focusing on visual explanations. However, none of these works investigated the bias that these explanations may contain and present to the user without further notice or discussion of the possible present bias.
As the explainability outcome is used to direct the design choices of algorithm developers as well as the decisions of system operators, the existence of bias in these explanations might lead to a chain of biased decisions.

In this work, we elaborate on the question "Are explainability tools gender-biased?" by performing a case study on face presentation attack detection (face PAD). Face PAD refers to an attack, in which an attacker attempts to impersonate another identity while using an FR system by, for example, using a video, a print, or a 3D mask. This task was chosen as it represents a simple binary classification task, which eliminates the influence of higher complexities. We investigate the bias in explainability outcomes of face PAD in terms of gender bias as it is one of the most well-known and discussed biases. For the case study, we utilize an arbitrary face presentation attack detector \cite{DBLP:journals/corr/abs-2209-09035} and two different, widely used explainability tools, GradCAM \cite{DBLP:conf/iccv/SelvarajuCDVPB17} and GradCAM++ \cite{DBLP:conf/wacv/ChattopadhyaySH18} and evaluate their explanations based on the presented gender using an deletion-and-insertion evaluation scheme \cite{DBLP:conf/bmvc/PetsiukDS18}. This is therefore the first work to investigate demographic bias in the outcomes of explainability tools.

\section{OUR CASE STUDY} 
\label{sec:case}

\begin{figure}
    \centering
    \includegraphics[width=1\linewidth]{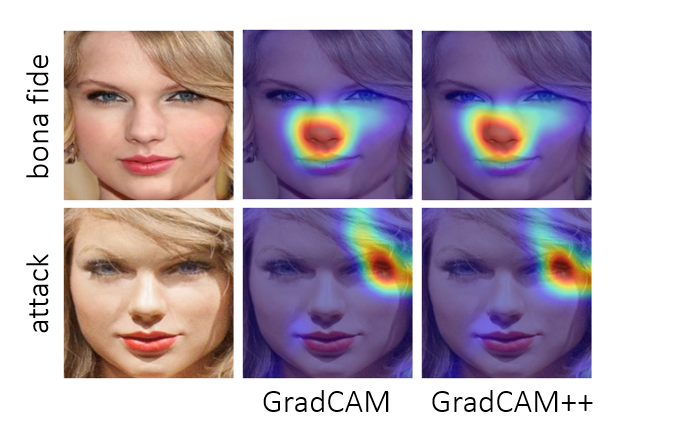}
    \caption{Example of explanation maps generated using GradCAM and GradCAM++ on a bona fide and an attack image. Both methods highlight a similar area in the image.}
    \label{fig:taylor}
\end{figure}

In this case study, we investigate the question if explainability tools provide explanations that are gender biased based on face PAD systems. These face PAD solutions are commonly trained to solve the PAD problem using binary classification between bona fide and presentations attack images. Recent works in the literature have demonstrated that state-of-the-art PADs are gender biased \cite{DBLP:journals/corr/abs-2209-09035}. In our experiments, we utilize three different models, one trained on a balanced gender dataset, and two models solely trained on male or female data. We then apply two explainability tools to generate visual explanations of the models' decisions. These explanations are then evaluated statistically based on insertion and deletion evaluation curves \cite{DBLP:conf/bmvc/PetsiukDS18}. By inserting and deleting the important pixels as identified by the explainability tool, we quantify the explainability performance - and performance differences based on gender, which would indicate gender bias presented in the explanations.

\subsection{Experimental Setup} 
\label{sec:experiments}
For the experiments we utilize three different models $PAD_{B}$, $PAD_{F}$, and $PAD_{M}$. All models share the same ResNet-50 \cite{DBLP:conf/cvpr/HeZRS16} architecture. We chose this architecture as it serves as a backbone in many state-of-the-art PAD methods \cite{DBLP:conf/eccv/ZhangYLYYSL20, DBLP:journals/tbbis/YuLSXZ21} and achieved good PAD performance \cite{DBLP:journals/corr/abs-2209-09035}. The $PAD_{B}$ model is trained on the training set of the CAAD-PAD dataset \cite{DBLP:journals/corr/abs-2209-09035} on images of both, males and females. We test on the testing split of the CAAD-PAD \cite{DBLP:journals/corr/abs-2209-09035} dataset, which provides a testing dataset consisting only of female or male images. The testing data consists of 53.827 male and 19.042 female images. The $PAD_{M}$ and the $PAD_{F}$ models are trained only on the male and female training sets of the CAAD-PAD dataset, respectively. We follow the implementation details provided in \cite{DBLP:journals/corr/abs-2209-09035}. Using models trained on different gender-based subsets allows us to investigate explanations obtained from models with different levels of bias. The used $PAD_{B}$ model achieved an Equal Error Rate (EER) of 2.54\% on the male test set and 3.00\% on the female test set. The $PAD_{F}$ and the $PAD_{M}$ achieved an EER of 2.96\% and 13.13\% on the male set and an EER of 5.90\%  and 10.62\% on the female test set, respectively, and thus clearly show bias as discussed in \cite{DBLP:journals/corr/abs-2209-09035}. 

As explainability tools, we use GradCAM \cite{DBLP:conf/iccv/SelvarajuCDVPB17} and GradCAM++ \cite{DBLP:conf/wacv/ChattopadhyaySH18} in our experiments. Both approaches produce saliency maps that highlight the important regions in an image for the predicted value. Examples of explanation maps produced by GradCAM and GradCAM++ are provided in Figure \ref{fig:taylor}.

\subsection{Evaluation Metrics} 
\label{sec:metrics}

\begin{figure}
    \centering
    \includegraphics[width=1\linewidth]{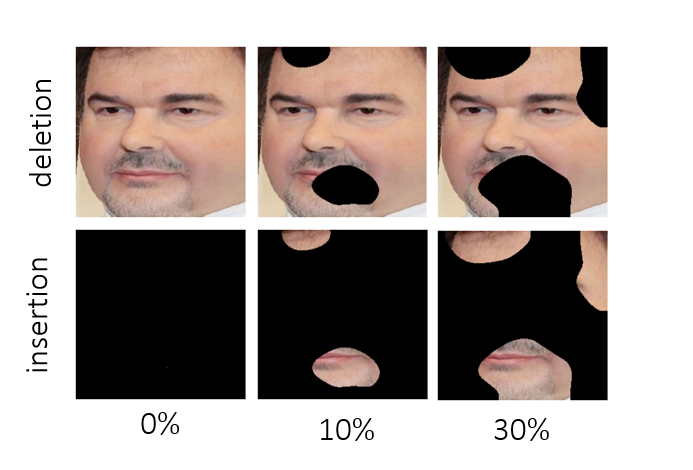}
    \caption{Visualisation of the insertion and deletion. Based on the calculated explanation map a fraction of the most important pixels are either removed or inserted.}
    \label{fig:evalscheme}
\end{figure}

To measure the bias of the explanations produced by the explainability tools, we utilize an insertion and deletion curve evaluation, following the trend in evaluating the performance of explainability outcomes \cite{DBLP:conf/bmvc/PetsiukDS18,9909844}. In the insertion evaluation, we iteratively insert pixels from the input image into a black canvas. The pixels, in this case, are selected based on the importance scores produced by the explainability tool. In the deletion evaluation, we iteratively delete pixels from the input image based on the calculated explanation map by setting their values to zero. After inserting or removing a certain amount of pixels based on their assigned importance to the decision, we evaluate the performance of the models on these newly generated images with the identified important pixels inserted or removed. In our experiments, we limited the amount of removed or added pixels, starting from 5\% up to 30\% with steps of 5\%, as the explanation maps often only indicate a small area as important to the decision. A visualization of the insertion and deletion evaluation procedure is shown in Figure \ref{fig:evalscheme}.

If the explainability methods do not produce gender bias, a similar performance should be observable, i.e. the accuracy in selecting the most important parts of the image to make the decision is similar for both male and female samples. 
If this is not the case, the explainability tools provide explanations that are gender biased.

\begin{figure*}
    \centering
    \begin{subfigure}{ %
     \includegraphics[width=0.23\textwidth]{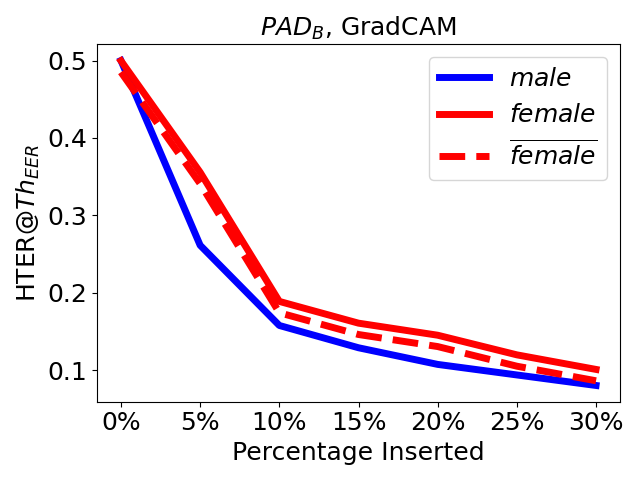}}
    \end{subfigure}
    \begin{subfigure}{ %
     \includegraphics[width=0.23\textwidth]{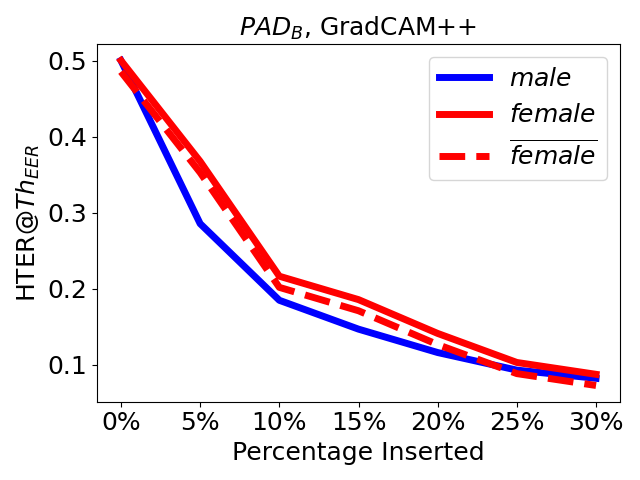}}
    \end{subfigure}
    \begin{subfigure}{ %
     \includegraphics[width=0.23\textwidth]{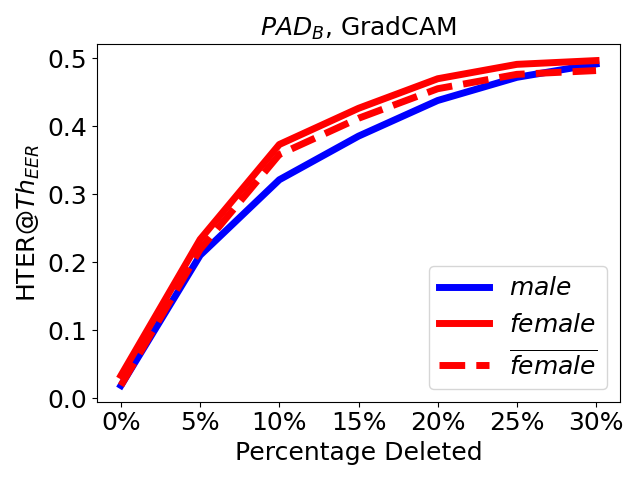}}
    \end{subfigure}
    \begin{subfigure}{ %
     \includegraphics[width=0.23\textwidth]{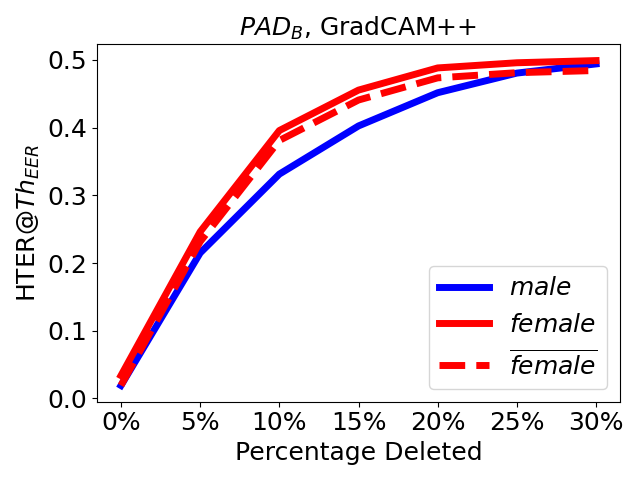}}
    \end{subfigure}
    \begin{subfigure}{ %
     \includegraphics[width=0.23\textwidth]{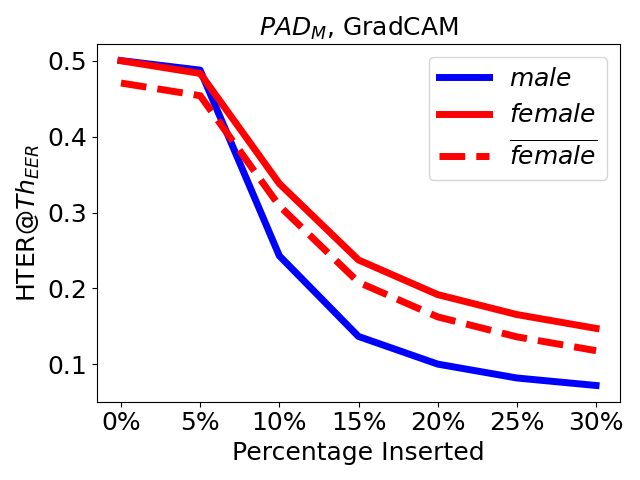}}
    \end{subfigure}
    \begin{subfigure}{ %
     \includegraphics[width=0.23\textwidth]{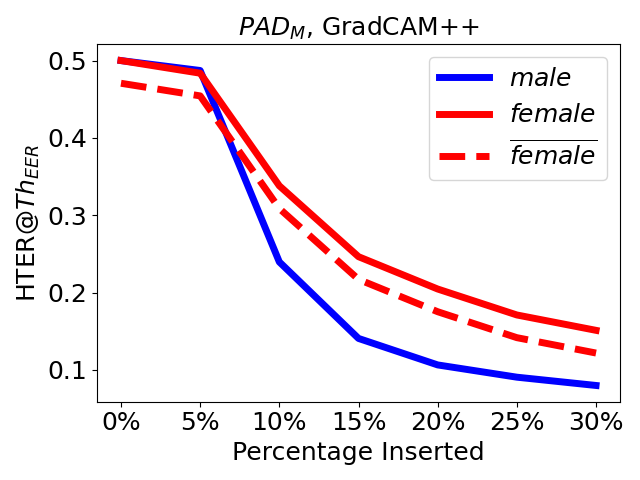}}
    \end{subfigure}
    \begin{subfigure}{ %
     \includegraphics[width=0.23\textwidth]{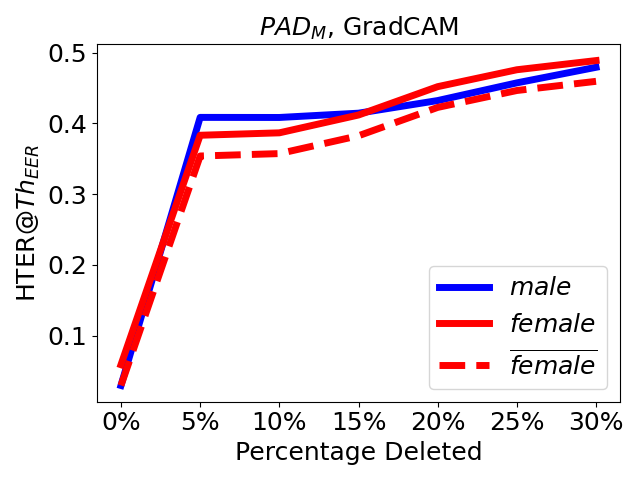}}
    \end{subfigure}
    \begin{subfigure}{ %
     \includegraphics[width=0.23\textwidth]{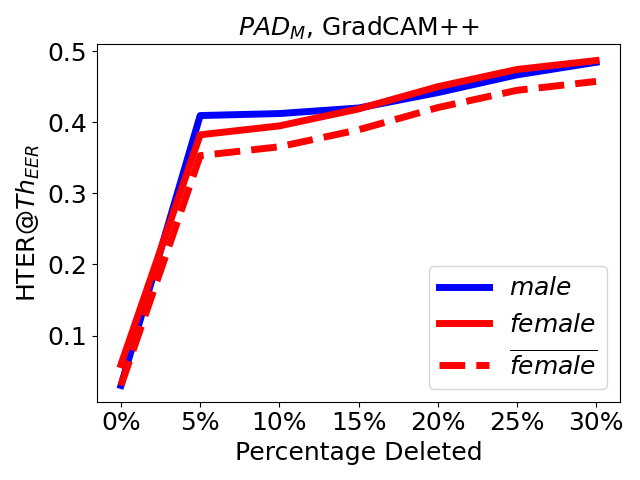}}
    \end{subfigure}
    \begin{subfigure}{ %
     \includegraphics[width=0.23\textwidth]{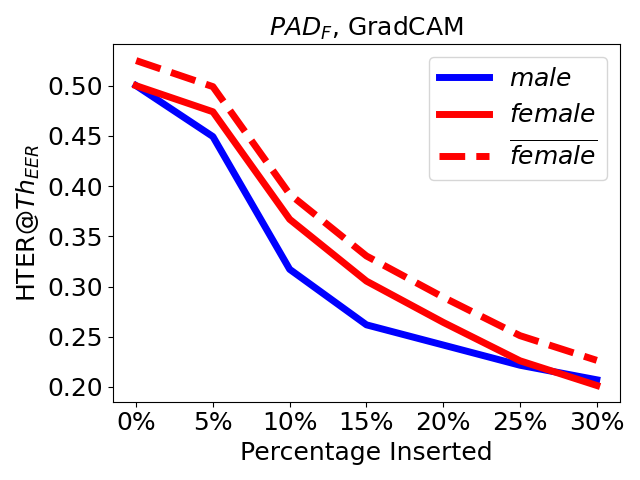}}
    \end{subfigure}
    \begin{subfigure}{ %
     \includegraphics[width=0.23\textwidth]{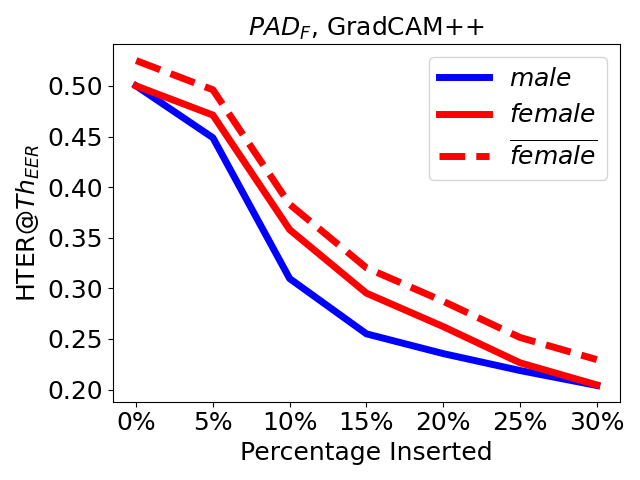}}
    \end{subfigure}
    \begin{subfigure}{ %
     \includegraphics[width=0.23\textwidth]{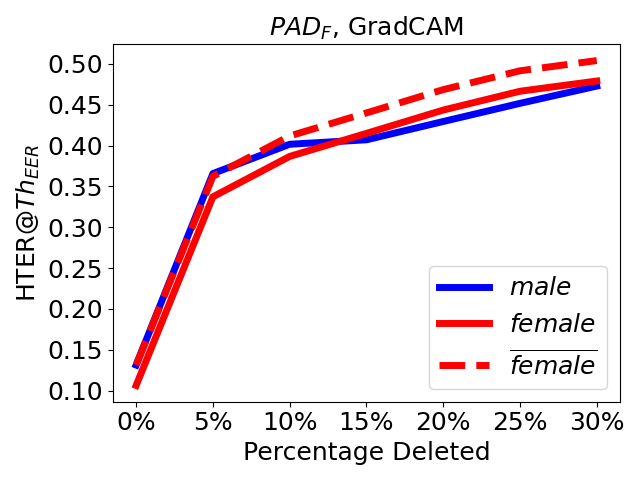}}
    \end{subfigure}
    \begin{subfigure}{ %
     \includegraphics[width=0.23\textwidth]{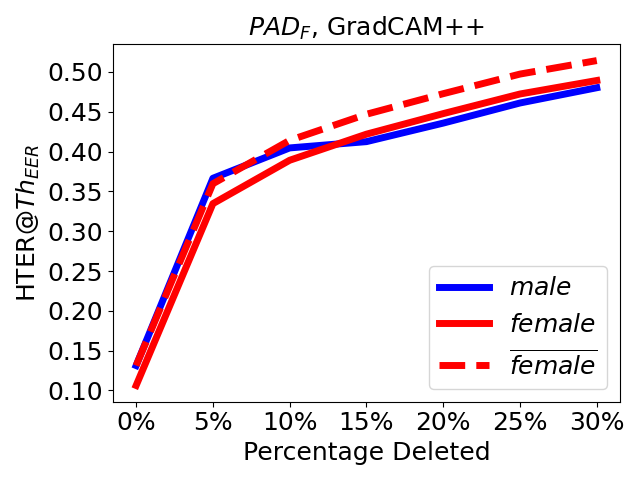}}
    \end{subfigure}
    \caption{Insertion and Deletion Curves for the three models, $PAD_{B}$ (balanced), $PAD_{M}$ (male), $PAD_{F}$ (female) using GradCAM \cite{DBLP:conf/iccv/SelvarajuCDVPB17} and GradCAM++ \cite{DBLP:conf/wacv/ChattopadhyaySH18}. The red-dotted line ($\overline{female}$) shows the normalized female performance with the same starting point as the male performance. Especially in the deletion evaluation curve, the gender bias of the model trained only on one gender ($PAD_{M}$ and $PAD_{F}$) can be observed as there is a performance difference between the error of evaluation of the male (blue) and female (red) dataset.}
    \label{fig:curves}
\end{figure*}

For the evaluation of the PAD system, we report the Half Total Error Rate (HTER) at the fixed threshold of the EER on the unaltered images. The HTER, which is widely used to report PAD performance \cite{DBLP:conf/wacv/FangDKK22, DBLP:conf/aaai/ShaoLY20, DBLP:journals/corr/abs-2303-02660}, is the half of the sum of the Attack Presentation Classification Error Rate (APCER) and the Bona Fide Presentation Classification Error Rate (BPCER) and allows us to report APCER and BPCER in a single curve. We also keep the threshold fixed for the evaluation of the different degrees of insertion or deletion as the threshold is also fixed in practice.

As the different models do not perform similarly on genders, we normalize the insertion and deletion curves with respect to the performance without manipulated images to provide comparable evaluation results. This allows a better visual interpretation and also allows us to calculate and report the Area-Under-the-Curve (AUC) as a quantified metric. For the normalization, we calculated the error of the models on the male and female data separately and then subtracted this initial performance difference from all following female error rates to get the normalized female performance. If the explanations of the models are not biased, the difference in the AUC should be zero, as the explanations would indicate with the same performance the most important pixels, independently of the gender of the presented sample.

\section{RESULTS}
\label{sec:results}
This section presents the results of our case study on face PAD using the considered explainability tools to investigate if the explanations provided are gender biased. First, we will look at the insertion and deletion curves of the different considered PAD models. Then we will quantify the gender bias in the explanations by comparing the AUC.

The results of the insertion and deletion evaluation are presented in Figure \ref{fig:curves}. It shows the insertion and deletion curves for all three models ($PAD_{B}$, $PAD_{M}$, and $PAD_{F}$) for both explainability methods. 
The red-dotted line ($\overline{female}$) in Figure \ref{fig:curves} indicates the normalized female performance to compensate different starting performance, as explained in Section \ref{sec:metrics}. This is needed to provide a fairer explainability comparison, as for example, the performance of the $PAD_{B}$ model on the female testing data is worse than on the male testing data, which should not influence the explanation performance comparison. 

The curves in Figure \ref{fig:curves} show the HTER  at the threshold of the EER (y-axis) over insertion or deletion proportion (x-axis). In the insertion curves, the error is decreased over higher insertion rate and vice versa for the deletion curves, i.e., the error is increased over an increased deletion rate.
In the insertion curve, a fast decrease indicates a better performance of the explainability tool, and a low AUC, therefore, indicates superior explainability performance. For the deletion curve, it is the other way around where a fast increase and a higher AUC indicate better performance of the explainability tool. The bias, therefore, is present if a performance gap is observable between explaining samples of different demographic groups. Therefore, the degree of bias can be indicated by the explainability performance difference when processing different groups.

The performance gaps (in term of HTER) over insertion and deletion rates (the outcomes of the explanation tools) between the case when the model is evaluated on male and female subsets, respectively, is observable in the insertion and deletion curves (Figure \ref{fig:curves}), indicating bias in the explanation outcome.

In the insertion curve, we observe bias as the decrease of the error when evaluated on male data is faster than the decrease in the error evaluated on the female dataset. 
This remains true even when normalizing the curve depending on the starting performance (dotted red line). Similar behavior can be observed on the more biased models, $PAD_{M}$ and $PAD_{F}$.
However, the bias in the explanations was smaller ($PAD_{B}$) than the bias present in the explanations of the $PAD_{M}$ model, at least by using GradCAM, as shown by the closer red and blue curves in Figure \ref{fig:curves}. 

Gender bias in explainability tools can also be observed in the deletion curves. In the $PAD_{B}$ model, the increase in error tends to be faster for females than for males. 
On the deletion curve for the $PAD_{M}$, we can observe a steeper increase in error as pixels are removed on the male data than in comparison to the female data, which indicates a gender bias. The opposite is true for the $PAD_{F}$ model, in which the error of the normalized deletion curve for the female testing data increases faster than the error on the male testing data, also indicating bias.

In addition to the insertion and deletion curves, we provide a quantified evaluation of the bias in explainability tools by reporting the AUCs and the performance difference between the male curve and the normalized female curve of each plot in Figure \ref{fig:curves}.


From the quantified results in Table \ref{tab:auc}, we made the following observations: a) The explainability tools are less biased when evaluated on the less biased $PAD_{B}$ model, than when they are evaluated on more biased PAD models ($PAD_{F}$ and $PAD_{M}$), b) in the explanations of the models $PAD_{M}$ and $PAD_{F}$, a higher performance difference is observable, indicating larger explainability gender bias. These observations are clearly observable for both explainability tools with slightly higher values for the GradCAM++ method and they are complementary for the ones reported early in this section based on the reported curves.

To conclude, we observed that the explainability tools are gender biased when they are used to explain the behavior of the considered PAD solutions. The bias in the explainability tools is, to some degree, lower for the PAD model that is less biased (trained on data of both genders) than when the explainability tools are evaluated on more biased PAD models (trained on gender-biased datasets). This might be due to bias that is present in the model investigated and inherits its bias to the explainability method. However, the exact roots of the bias in the explainability performance have to be further investigated. Interestingly, we also noticed a link between the bias in the PAD models and the bias in the explanation. In the deletion curves for the $PAD_{M}$ and $PAD_{F}$, the bias manifests in the same direction as the models' bias. The performance on the male samples is better than the performance on the female samples using the $PAD_{M}$ model that has been trained solely on male images. The same is the case for the $PAD_{F}$ with better performance for female images, while it was trained on female images.

\section{CONCLUSION}
\label{sec:conclusion}
In this work, we investigated the research question: "Are explainability tools gender-biased?", while taking the explanation of face PAD behavior as an example. 
In our effort to answer this question, we performed a case study on the problem of face PAD by using two explainability tools, GradCAM and GradCAM++, and PAD models with different levels of gender bias. Our investigation concluded that there are differences in the explainability performance when explaining male and female samples and thus there is gender bias in the explainability outcome. As the explainability outcomes are used to increase the transparency for developers and system operators, the existence of bias in these explanations is of concern and needs attention. Future research works could investigate whether other bias factors, such as ethnicity and age, or even non-demographic biases are also affecting explainability outcomes, along with investigating the bias-inducing factors and bias mitigation possibilities.

\begin{table}[]
\centering
\resizebox{\linewidth}{!}{
\begin{tabular}{llllllll}
 & \multicolumn{3}{l}{Evaluation Data} &  & \multicolumn{3}{l}{Evaluation Data} \\ \hline \hline
Deletion & Male & Female & $\Delta$ & Insertion & Male & Female & $\Delta$ \\ \hline
GradCAM &  &  &  & GradCAM &  &  &  \\ \hline
$PAD_{B}$ & 0.104 & 0.109 & \textbf{0.005} & $PAD_{B}$ & 0.052 & 0.059 & \textbf{0.007} \\
$PAD_{M}$ & 0.119 & 0.110 & 0.009 & $PAD_{M}$ & 0.067 & 0.078 & 0.011 \\
$PAD_{F}$ & 0.118 & 0.125 & 0.007 & $PAD_{F}$ & 0.092 & 0.107 & 0.015 \\ \hline
GradCAM++ &  &  &  & GradCAM++ &  &  &  \\ \hline
$PAD_{B}$ & 0.107 & 0.113 & \textbf{0.006} & $PAD_{B}$ & 0.056 & 0.061 & \textbf{0.005} \\
$PAD_{M}$ & 0.120 & 0.111 & 0.009 & $PAD_{M}$ & 0.068 & 0.080 & 0.012 \\
$PAD_{F}$ & 0.119 & 0.126 & 0.007 & $PAD_{F}$ & 0.091 & 0.106 & 0.015 \\ \hline
\end{tabular}}
\vspace{1.2mm}
\caption{\textbf{AUC for the different PADs, evaluation data genders, and explainability methods for both curves}:
The higher difference between AUC of male and female indicates higher bias in the explainability performance, the lowest difference (bias) is in bold. As for the bias in the PAD performance of the $PAD_{B}$, the bias in its explanation is lower than the other PADs. 
}
\label{tab:auc}
\end{table}


\bibliographystyle{IEEEtran}
\bibliography{conference_101719}

\end{document}